# BAFPN: Bidirectionally Aligning Features to Improve Object Localization Accuracy in Remote Sensing Images

Jiakun Li[a], Qingqing Wang[a], Hongbin Dong*[a], Kexin Li[b]

[a]: Harbin Engineering University, Heilongjiang, China
[b]: Shandong Jianzhu University, Shandong, China
*: Corresponding author at: Harbin Engineering University, Heilongjiang, China
E-mail address: donghongbin@hrbeu.edu.cn

*Abstract*—Current advanced visual model detectors typically use feature pyramids to extract multi-scale information, with Feature Pyramid Network (FPN) being the most commonly used classic architecture. However, the classic FPN and its variants (AUGFPN, PAFPN, etc.) have not fully considered spatial misalignment at the global scale, resulting in insufficient high-precision localization performance for target detection. This paper proposes a novel Bidirectional Feature Alignment Feature Pyramid Network (BAFPN), which aligns the position and edges of the target at the global scale through the Spatial Feature Alignment Module (SPAM) in the prepositive bottom-up information propagation path and supplements precise target location information to the deep features. It then performs top-down semantic alignment through the Fine-grained Semantic Alignment Module (SEAM) to alleviate the aliasing effect generated during cross-scale feature fusion. In addition, we replace the 1×1 lateral connections with the Group Aggregation Lateral Connection Module (GALM) to retain the rich channel information in the raw features output by the Backbone. On the DOTAv1.5 dataset, BAFPN improves the AP75 of the baseline model by 1.68 percentage points, and the AP50 and mAP are improved by 1.45 and 1.34 percentage points, respectively. BAFPN also significantly improves the performance of several other advanced detectors.

## I. Introduction

Object detection is a fundamental task in the field of computer vision and is widely applied in industrial scenarios such as object tracking and pedestrian detection. The detection of objects in remote sensing images presents a higher level of challenge due to the drastic variation in object scales, dense arrangements, and the large number of small objects. Many efforts have been made to improve various components of traditional object detectors to enhance their performance in detecting objects in remote sensing images.

In remote sensing images, the drastic variation in object scales often requires sufficient spatial detail information to assist the model in accurately localizing the objects in the image. Feature Pyramid Networks (FPN) [1], with their low computational cost and ability to fuse deep semantic information and shallow object location information, have been widely applied in both single-stage and two-stage object detectors.

FPN uses 1×1 convolutions to reduce the number of channels in features output by the Backbone across different layers to a unified smaller value, which causes information loss. Info FPN [2] leverages the Pixel Shuffle operation from the field of image super-resolution to transform channel information into spatial information and aggregates features at the pixel scale, thereby reducing information loss. CE-FPN [3] proposes a sub-pixel skip fusion operation based on Pixel Shuffle to replace the original 1×1 convolution and linear upsampling. This operation extracts a feature layer containing enough channels, and converts the deep features into spatial features through the Pixel Shuffle operation, which are then directly fused into a shallower feature layer, thereby completely retaining the information in the original feature map. FaPN [4] constructs a feature selection module through an attention mechanism, weighting the channels before reducing their number to retain more critical information. t pixel-levelA2-FPN [6] extracts and distributes additional information from various levels to compensate for the information loss caused by channel compression.

To address the issue of feature misalignment during multi-level feature map fusion, Li et al. [7] use a method similar to optical flow alignment, where a convolutional layer learns feature offsets between adjacent feature maps and generates an optical flow field, and deep features are sampled based on the flow field to align the two feature maps. FaPN [4] generates a more refined displacement field using a similar approach and applies deformable convolutions to adjust the spatial locations of deep feature map spatial features. InfoFPN [2] uses a template matching mechanism, computing multiple templates to be matched based on shallow feature maps using a convolutional network, then calculating the probability of matching between deep feature maps and each template, and finally selecting the best template to combine with the shallow feature map to generate a displacement field for aligning deep feature maps.

Despite the excellent performance demonstrated by the aforementioned works across multiple datasets, they have not addressed the problem of feature misalignment at the global scale. When the original image undergoes feature extraction by the backbone network and outputs deep features, it often passes through dozens or even hundreds of convolution operations. During this process, the objects in the image may experience positional shifts and shape distortions, which affect the detector's ability to accurately localize the objects. In the top-down feature cross-scale fusion process of FPN, the shifted and distorted object features will be fused at incorrect positions, activating irrelevant shallow features and ultimately leading to suboptimal detector performance. Simply aligning adjacent feature maps cannot improve the misalignment issue in deep features relative to the original image; it is necessary to align features at the global scale to ensure that the position and shape of the objects in the deep features remain consistent with those in the original image. To address this, we propose a novel BAFPN. BAFPN aligns features at the global scale through a spatial feature alignment module combined with a bottom-up information propagation path before the top-down feature cross-scale fusion, while the Spatial-to-Depth down-sampling module supplements high-precision target location information to the deep features, further enhancing the model's target localization ability. To improve the aliasing effect that occurs during multi-layer feature fusion without losing the diversity of feature representations, we propose a lightweight Fine-Grained Semantic Alignment Module. This module generates global masks applied to each channel-pixel of the feature map by learning the global contextual differences and fine-grained pixel-level differences between adjacent feature layer channels. This module can reduce the semantic gap between features while preserving the diversity of feature representations. To reduce the information loss caused by channel reduction by 1×1 convolution in FPN, we designed a Group-Aggregate Lateral Connection Module, which semantically groups channels through convolutions with shared parameters and aggregates different semantic groups using multiple sets of learnable weights, thereby generating low-channel features with rich semantic information.

Our work can be summarized as follows:

- A novel FPN architecture: It includes a prepositive bottom-up information supplementation path and a top-down feature cross-scale fusion path.
- Spatial feature alignment at the global scale: Through the prepositive bottom-up information supplementation path and SAPM, features with positional shifts and shape distortions are aligned at the global scale.
- Fine-grained semantic alignment: By generating global masks using SEAM, the semantic gap between features is reduced while preserving the diversity of feature representations.
- Group aggregation lateral connection: By grouping the channels in the features according to semantics and then aggregating them, a large amount of information loss can be avoided.

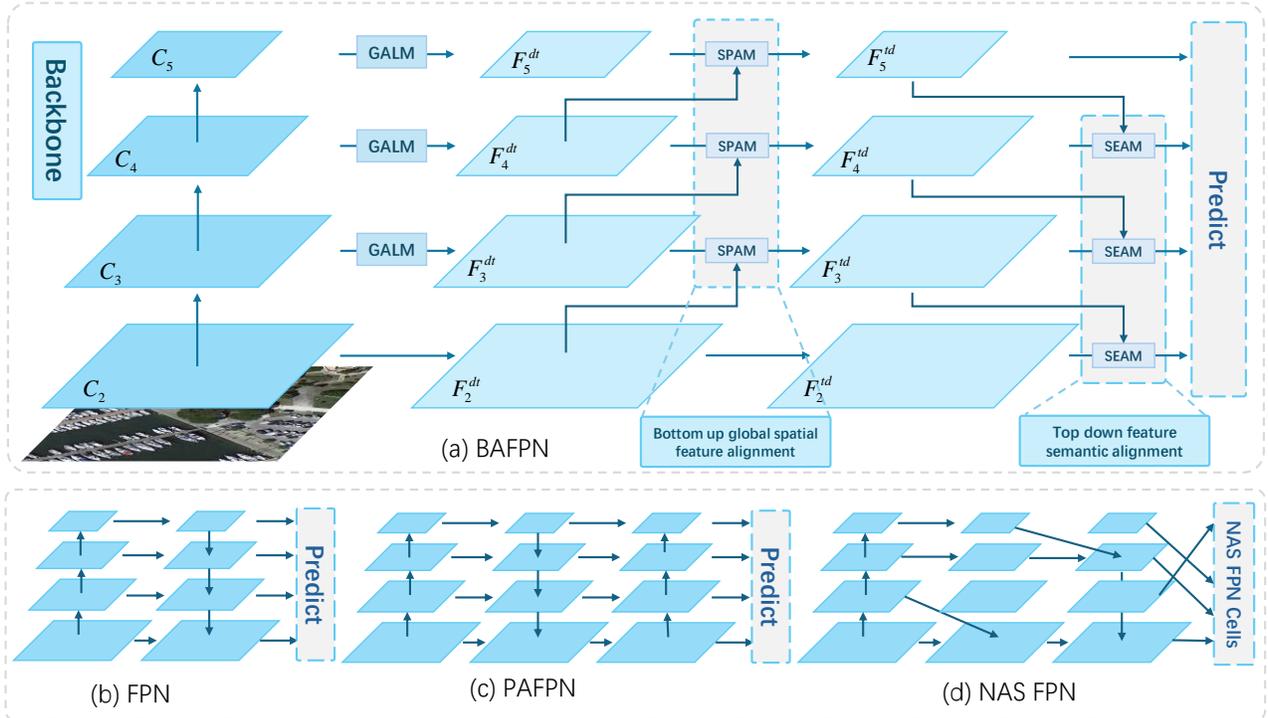

Fig. 1. The overall structure of BAFPN and other FPNs

## II. RELATED WORK

### A. Rotated Object Detection

Oriented object detection aims to recognize and locate objects rotated in arbitrary directions, which is particularly suitable for object detection in aerial and remote sensing images. Traditional object detection methods typically assume that objects are axis-aligned rectangles, which are not ideal for detecting directional objects such as text or objects in remote sensing images. To address this issue, Ma et al. proposed a Rotated Region Proposal Network [8] that can generate skewed proposal boxes with angle information, using this angle information for bounding box regression to achieve directed object detection. Yang et al. proposed an object detection method called H2RBox [9], which uses only the annotations of horizontal boxes for weakly supervised training, achieving performance comparable to methods that use rotated boxes and reducing the annotation cost of rotated boxes. Qian et al. proposed a Unified Transformation Strategy (UTS) [10], which converts the bounding box regression loss in horizontal object detection to directed object detection. UTS, combined with the rotated intersection-over-union loss (RIoU), facilitates the transfer of horizontal bounding box regression to directed bounding box regression.

Due to the periodic nature of angles, the intersection-over-union (IoU) of rotated boxes and critical angle changes can lead to abrupt loss variations. To solve this issue, Ming and Yang et al. [11][12] discretized the rotation angle and transformed the angle regression task into a classification task, thereby avoiding loss discontinuities caused by the periodicity of angles. GWD [13] treats the rectangular box as a two-dimensional Gaussian distribution and uses the Gaussian Wasserstein distance as a regression loss function, providing a more streamlined solution to this problem. Qian et al [14] introduced 8 regression parameters and modulation rotation loss to avoid rotational sensitivity errors. TricubeNet [15] uses heatmap regression to direct the target, representing each target as a 2D Tricube kernel, replacing traditional directed box offset regression, and solving the angle continuity issue.

To address the high-quality feature extraction problem for multi-directional objects in oriented object detection, Pu et al. proposed an Adaptive Rotated Convolution Module (ARC) [16], which can adaptively adjust the angle of convolution kernels based on the target in the input image and introduces an effective conditional computation mechanism to accommodate large directional changes of objects in the image. LSKNet [17] uses a series of large-kernel depthwise separable convolutions to dynamically adjust large spatial receptive fields according to the different needs of objects, thereby better modeling the contextual scope of various objects in remote sensing scenes.

### B. Multi-Scale Feature Learning

In remote sensing images, the target size differs significantly, and while feature detection based on a single scale can achieve faster detection speeds, it fails to accurately detect targets at different scales. Using image pyramids to construct feature pyramids and performing independent computations on features at each scale is effective but incurs significant computational overhead. FPN reduces the cost of independent calculations at each feature layer by adjusting the feature channels to a uniform smaller value through lateral connections and constructs a top-down path for propagating semantic information from high-level features, thereby activating low-level features. This method is fast and accurate, but its shortcomings are also quite prominent.

Liu et al. argued that deep features lose some spatial detail information compared to shallow features. The Path Aggregation Network (PAN) [18] adds a bottom-up feature propagation branch after FPN, shortening the path for propagating high-resolution spatial features from shallow layers to deep feature maps, thereby achieving bottom-up path enhancement. $A^2FPN$ [19] extracts global contextual information from all feature layers and allocates it to each feature layer to improve recognition accuracy. CFPN [20] also adopted a similar idea, aggregating multi-scale features from different levels and distributing the aggregated features across all relevant layers, so that each layer of features contains both semantic and significant detail information. LrFPN [21] extracts refined location information only from the bottom-most feature map and supplements it to all other feature maps. Park et al. proposed a scale-sequence feature-based feature pyramid network, ssFPN [22], which enhances high-resolution feature maps by performing 3D convolutions along the hierarchical axis of FPN to improve small target detection performance. Recursive FPN [23] applies a "look twice" mechanism in computer vision by feeding the output of the FPN fusion back into the backbone for a second cycle, achieving outstanding performance across multiple datasets. NAS-FPN [24] and AutoFPN [25] search for the optimal FPN architecture through network architecture search. AFPN [40] progressively fuses non-adjacent feature layers, reducing information loss while avoiding aliasing effects caused by semantic gaps.

Gong et al. addressed the limitations of FPN in small target detection by proposing a fusion factor estimated through statistical methods [26], which controls the information transfer from deep to shallow layers to adapt FPN for small target detection. Xiang et al. proposed a retrospective feature pyramid network, Retro-FPN [27], which innovatively introduced a retro-transformer [41], and effectively extracts semantic features for each point through explicit and retrospective feature refinement processes.

### III. METHOD

#### A. Grouped Aggregation Lateral Connection Module

The deep-level features output by the backbone typically have a high number of channels. Traditional lateral connections usually compress the number of channels in feature maps from different levels to a unified, smaller value using a single 1×1 convolution layer. This approach directly fuses numerous channels with different semantics, which is not only inefficient but also leads to information loss, compromising the completeness of feature representation.

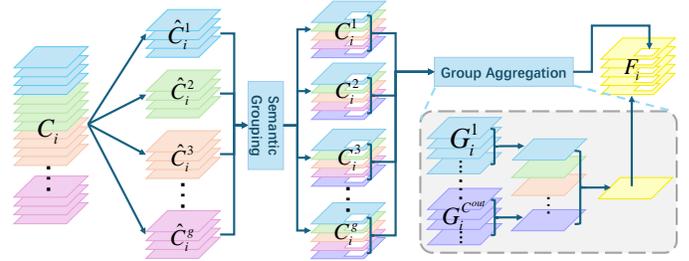

**Fig. 2. Group Aggregation Lateral Model**

In fact, the features output by the backbone network often contain many channels with similar semantics. If channels with similar semantics are grouped together first and then weighted and fused with channels of different semantics, the fusion efficiency can be significantly improved, and richer feature representations can be obtained. To this end, we propose a Grouped Aggregation Lateral Connection Module (GALM). This module divides the channels of the feature map into $g$ groups, where each group learns a consistent channel semantic distribution using convolution kernels with shared parameters. Then, channels with similar semantics are categorized into the same semantic group. Finally, features of all semantic groups are aggregated using multiple sets of learnable weights. This approach effectively preserves the rich channel information of the original features while introducing only a small number of additional parameters. The overall structure of GALM is shown in Figure 2. Specifically, GALM divides the *i*-th layer feature $C_i$ output by the backbone network into $g$ groups, expressed as: $C_i = C_i^1 \circ C_i^2 \circ ... \circ C_i^g$ where ∘ denotes the concatenation operation. Each group of features learns a unified channel semantic distribution via a shared 1×1 convolution kernel [2], which can be expressed as:

$$C_i^j = \psi^1(\hat{C}_i^j)$$

$$C_i^j \in \mathbb{R}^{B \times \frac{C^{in}}{g} \times H \times W}, \hat{C}_i^j \in \mathbb{R}^{B \times C^{out} \times H \times W}$$

$$C_i^j = C_i^{j,1} \circ C_i^{j,2} \circ ... \circ C_i^{j,C^{out}}$$

Where $\psi^k$ represents a convolution kernel with a size of $k$. The semantic group is defined as $G_i^t = \{C_i^{1,t}, C_i^{2,t}, ..., C_i^{g,t}\}, (1 \leq t \leq C^{out})$. Since the feature channels within the same group share the same convolution kernel parameters, they exhibit similar semantics. Subsequently, GALM performs weighted fusion of features from different semantic groups using $C^{out}$ sets of learnable weights, thereby generating a richer lateral connection output while simultaneously adjusting the number of feature channels:

$$F_{i,s} = \sum_{t=1}^{C^{out}} \sum_{j=1}^{g} W_{i,s}^{j,t} \cdot C_i^{j,t}$$

where $F_{i,s}$ represents the *s*-th channel of the $F_i$, and $W_{i,s}^{j,t}$ is the learnable weight for channel $C_i^{j,t}$ in semantic group $G_i^t$, used to generate $F_{i,s}$.

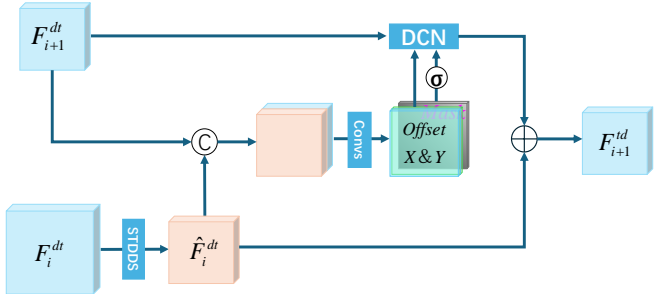

**Fig. 3. Spacial Features Align Module**

### B. Spatial Feature Alignment at Global Scale

The structure of the Spatial Feature Alignment Module (SPAM) is shown in Figure 3. Traditional FPN variants with feature alignment functions, such as FaPN [4] and FAFPN [7], align features only between adjacent feature maps. This approach can only address the feature misalignment problem during feature fusion and cannot improve the feature offset and shape distortion of deep features relative to the original image. To fundamentally resolve these issues, we introduce an additional bottom-up information propagation path and Spatial Feature Alignment Module (SPAM) before the top-down cross-scale feature fusion path in FPN. By combining the two, spatial features can be precisely aligned at the global scale, enhancing the consistency of target positions in different feature levels and significantly improving target localization accuracy.

SPAM uses our designed Spatial-to-Deepth down-sampling module (STDDS) for down-sampling. Compared with traditional pooling down-sampling or convolution down-sampling with a stride of 2, STDDS can reduce feature resolution while retaining and enhancing spatial detail information. After STDDS down-sampling, shallow feature will be concatenated with deep feature in the channel dimension, and the offset field and sampling mask of the deep feature relative to the shallow feature will be generated through the convolution layer. The offset field and sampling mask will be embedded in a deformable convolutional network (DCN) [28] to achieve precise alignment of deep features through a dynamic sampling mechanism based on the offset field. In order to make full use of the bottom-up information propagation path, the down-sampling result of shallow feature will be fused into the aligned deep feature, so that high-precision spatial position information can be quickly transmitted upward with low loss, thereby further enhancing the detector's positioning performance for the target.

*a. Spatial-to-Depth Down-Sampling (STDDS)*

Shallow feature maps contain more detailed information that helps locate the target, and their target positions are usually closer to the original image. This characteristic determines that the alignment of deep features requires the guidance of shallow features.

Before feature alignment, shallow feature must be down-sampled to match the scale of deep feature. Traditional maximum pooling down-sampling will lose some detailed information. Although average pooling can retain the original information to a certain extent, it is difficult to highlight salient features. Convolutional down-sampling with a stride of 2 will significantly increase the number of model parameters. brings additional computational burden. To preserve rich detail information in shallow feature while filtering noise, we propose a lightweight Spatial-to-Depth Down-Sampling Module (STDDS).

The structure of STDDS is shown in Figure 4. Firstly, STDDS weights the pixels in the feature map through spatial attention [29] to enhance important spatial regions, which is mathematically expressed as:

$$M_S^{F_i^{dt}} = \sigma(\psi^k(\frac{1}{C}\sum_{j=1}^{c} F_{i,j,a,b}^{dt} \circ \max_{j=1,...,C} F_{i,j,a,b}^{dt})) \cdot F_i^{dt}$$

Here, $F_{i,j,a,b}^{dt}$ represents the pixel value at coordinate (*a*, *b*) in the *j*-th channel of the feature map output by the *i*-th lateral connection module. $\sigma$ is the sigmoid activation function. Next, weighted features are sampled according to the following rules:

$$\begin{cases} F_i^{dt\,(00)} = F_{i\;:,:,0:H:2,0:W:2}^{dt} \\ F_i^{dt\,(01)} = F_{i\;:,:,0:H:2,1:W:2}^{dt} \\ F_i^{dt\,(10)} = F_{i\;:,:,1:H:2,0:W:2}^{dt} \\ F_i^{dt\,(11)} = F_{i\;:,:,1:H:2,1:W:2}^{dt} \end{cases}$$

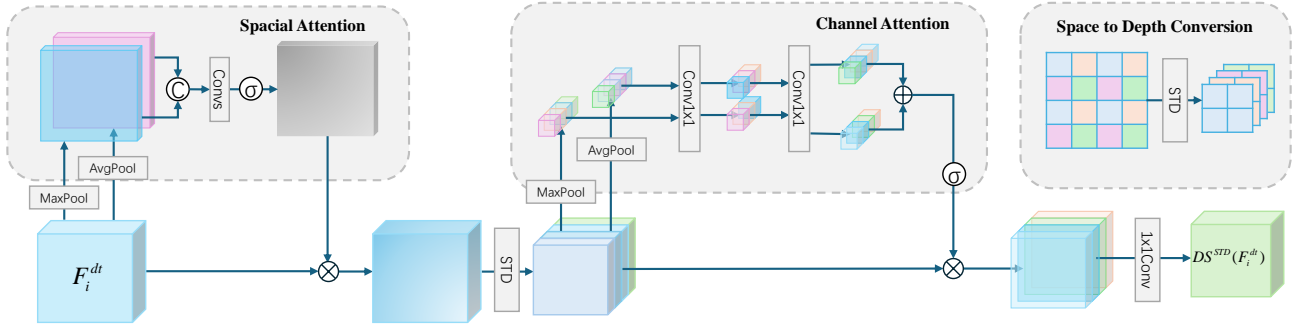

Fig. 4. Spacial to Depth Down Sample

These sub-sampling operations extract interleaved pixel blocks from the spatial dimensions of the input feature map. The four sub-sampled results are concatenated along the channel dimension to transform spatial information into depth information, described as:

$$STD(F_i^{dt}) = F_i^{dt(00)} \circ F_i^{dt(01)} \circ F_i^{dt(10)} \circ F_i^{dt(11)}$$

$$STD(F_i^{dt}) \in \mathbb{R}^{B \times 4C \times \frac{H}{2} \times \frac{W}{2}}$$

Subsequently, STDDS applies channel attention [29] to weight the concatenated feature channels, further enhancing spatial information at a larger scale. Finally, the number of feature channels is compressed to match the deep feature through a single-layer 1x1 convolution. The overall mathematical formulation of STDDS is:

$$\hat{F}_i^{dt} = DS^{STD}(F_i^{dt}) = \psi^1(M_C\{STD(M_S^{F_i^{dt}})\})$$

By transforming spatial information into depth information, STDDS enables the model to flexibly combine and process original features. At the same time, STDDS introduces spatial attention and channel attention mechanisms in the spatial to depth downsampling process, weighting spatial features in two stages at different scales, achieving the preservation and enhancement of spatial information in the original features while downsampling.

*b. Spatial Feature Alignment and Feature Fusion*

DCN introduces learnable offsets, enabling the convolutional kernel to adaptively adjust its shape and sampling positions during the convolution operation. This allows it to better accommodate the diverse object shapes in the input features. Consequently, SPAM employs DCN to align two adjacent feature maps.

The downsampled output from the STDDS module matches the scale of the deep feature while retaining high-precision location information from the shallow feature. This allows the shallow feature to directly guide the alignment of the deep feature and fuse with it. SPAM uses a combination of a 1×1 convolution and a 3×3 convolution to learn the spatial differences between the two feature maps. This process generates the offset field and the corresponding sampling weights for the deep features relative to the shallow features. Subsequently, DCN adjusts the spatial position of depth feature using offset fields and controls the contribution of different pixels to the convolution results through sampling weights, aligning them with shallow feature. The convolution result at any position for DCN can be expressed as:

$$\hat{x}_{i,j} = \sum_{p=-\lfloor \frac{k}{2} \rfloor}^{\lfloor \frac{k}{2} \rfloor} \sum_{q=-\lfloor \frac{k}{2} \rfloor}^{\lfloor \frac{k}{2} \rfloor} w_{p,q} \cdot x_{i+p+\Delta x, j+q+\Delta y} \cdot v_{i+p,j+q}$$

Where $w_{p,q}$ represents the convolution kernel weight at position $(p, q)$, $\Delta x$ and $\Delta y$ are the additional offsets embedded into DCN, and $v_{i+p,j+q}$ is the additional sampling weight applied to $w_{p,q}$ at image coordinate $(i, j)$. Finally, SPAM fuses the down-sampling result from STDDS with the aligned deep feature. The overall process of SPAM can be described as:

$$SPAM(F_i^{dt}, F_{i+1}^{dt}) = DCN^{\psi^{OM}(DS^{STD}(F_i^{dt}) \circ F_{i+1}^{dt})}(F_{i+1}^{dt})$$

Where $\psi^{OM}$ represents the convolutional layers used to generate the offsets and sampling weights, and $DCN^{OM}(F)$ denotes the deformable convolution operation on feature map $F$, guided by the offsets and sampling weights $OM$.

In terms of implementation SPAM only align two adjacent feature maps, but in BAFPN, the alignment process is performed from bottom to top, which allows the targets in each feature layer to adjust their spatial positions based on the features of the lowest layer, thereby solving the problems of feature position offset and shape distortion on a global scale.

*C. Fine-Grained Semantic Alignment*

Info-FPN [2] includes a semantic encoder composed of multiple layers of convolutions, which simultaneously encodes two feature maps during cross-scale feature fusion to reduce the semantic gap between the two layers of features, thereby alleviating the aliasing effect generated during cross-scale feature fusion. Li et al. [7] introduced a channel attention mechanism in FPN, which computes the channel weights for each feature layer based on a set of shared parameters and applies a unified weighting across all feature layers to guide them in learning unified semantics. Although this method can alleviate the aliasing effect, the unified coarse-grained weighting may suppress the diversity of feature representations.

Compared with FPN, BAFPN introduces additional positional information from lower-level features into each feature layer, resulting in more severe aliasing effects. To mitigate the aliasing effect in cross-scale fusion while preserving the diversity of feature representations, we propose a Channel-Pixel-level Fine-grained Semantic Alignment Module (SEAM). Figure 5 shows the overall structure of SEAM.

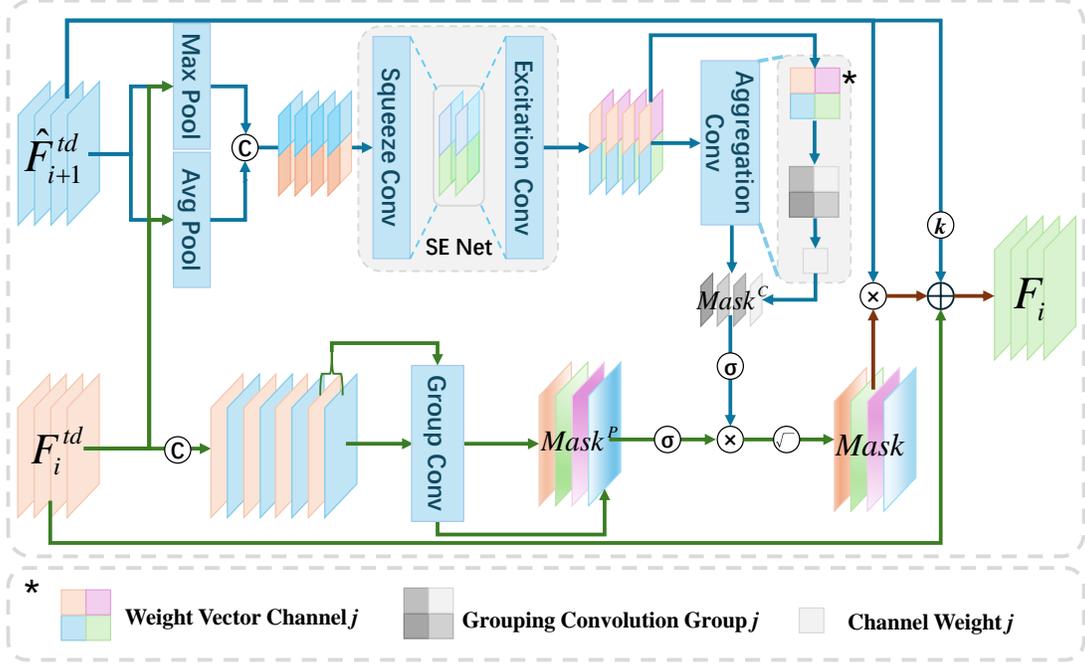

**Fig. 5. Semantic Align Module**

Specifically, SEAM consists of a global context-weighting branch and a pixel-weighting branch. The global context-weighting branch compresses the feature map along the channel dimension into four sets of global feature vectors through max-pooling and average-pooling operations. These global features are then concatenated along the pixel dimension and input into a Squeeze-Excitation network [30] composed of two 1×1 convolutions for coarse-grained channel weight modeling, producing four sets of weight modeling vectors. Subsequently, the four sets of weight vectors are aggregated at the channel scale via a single 2×2 grouped convolution, generating coarse-grained channel weights, which explicitly introduce the global interaction between the two feature maps. The mathematical expression of this process is as follows:

$$F = max^C(\hat{F}_{i+1}^{td}) \bullet max^C(F_i^{td}) \bullet avg^C(\hat{F}_{i+1}^{td}) \bullet avg^C(F_i^{td})$$

$$Mask^C = \sigma(\psi^{2,C^{out}}(SE(F)))$$

where $\psi^{k,s}$ is the convolution operation with size $k$ and group $s$, and $\bullet$ denotes the operation of combining vectors along the pixel dimension.

In the pixel-weighting branch, $\hat{F}_{i+1}^{td}$ and $F_i^{td}$ are interlaced along the channel dimension and input into a 7×7 convolution with $C^{out}$ groups. This branch learns the pixel-level feature differences between $\hat{C}_j \in \hat{F}_{i+1}^{td}$ and $C_j \in F_i^{td}$ through convolution groups $Conv_j$, generating fine-grained pixel-level weights, and further explicitly introduces the local interaction between the same channels of the two feature maps. This process can be expressed as:

$$Mask^P = \sigma(\psi^{7,C^{out}}(\hat{F}_{i+1}^{td} \otimes F_i^{td}))$$

Where $\otimes$ denotes the operation of interlacing tensors along the channel dimension.

Directly multiplying different masks will result in stronger suppression of features and hinder the flow of gradients. Therefore, after obtaining the product of the channel and pixel masks, we root it to balance the effects of different scale masks, resulting in the final global mask. Considering that additional residual connections during feature fusion might reduce the prominence of shallow feature, deep residual feature will also be multiplied by a learnable saliency weakening factor $k(1 \leq k \leq 0)$ before fusion. Overall, the two layers of features are fused as follows:

$$\hat{F}_{i+1}^{td'} = \sqrt{Mask^P \cdot Mask^C} \cdot \hat{F}_{i+1}^{td} + k \cdot \hat{F}_{i+1}^{td}$$

$$F_i = \hat{F}_{i+1}^{td'} + F_i^{td}$$

In the case where the number of input and output channels is the same, the number of parameters in SEAM is smaller than that of a single 1×1 convolution, thus SEAM remains lightweight. SEAM generates a global mask by learning the global context relationships and fine-grained spatial differences of the two feature maps, and applies channel-pixel-level weighting to the deep feature map before fusion. This allows the deep features to fully consider their similarities and differences with the shallow features, and adaptively adjust their feature representations, thus reducing the semantic gap between the two feature maps while maintaining the diversity of feature representations.

## IV. EXPERIMENTS

### A. Dataset

We validate the effectiveness of BAFPN in improving localization accuracy using the public dataset DOTAv1.5 [37]. DOTAv1.5 uses the same images as DOTAv1.0, but it additionally annotates extremely small instances smaller than 10 pixels and introduces the category "container crane," making it more challenging. Before training, the image sizes in the dataset are cropped into 1024x1024 sub-images, with an overlap of 200 pixels between adjacent sub-images.

The evaluation metrics in this paper are the commonly used AP50, AP75, and mAP in the field of object detection. Among them, mAP is the mean average precision at different IoU thresholds. AP75 is the average precision computed with an IoU threshold of 0.75, where a detected bounding box is considered a correct detection only if its overlap with the ground truth bounding box is 75% or higher. Compared to AP50, AP75 has stricter overlap requirements, and the detection model needs to localize the object more precisely.

### B. Experimental Setup

All experiments are conducted on a platform equipped with a single RTX4090 GPU, an Intel Core i7-13700K CPU, and 32GB of RAM, running on Windows 10. The Python version is 3.8.0, with PyTorch 2.0.0 and CUDA 11.8. All experiments are implemented using the mmrotate framework, and to ensure fairness, the hyperparameters and other settings of the models used for comparison are configured with the default settings of mmrotate.

TABLE I. CLASS-WISE ACCURACY COMPARISON (AP50)

RFR: ROTATED FASTER RCNN[31], RTS: ROI TRANSFORMER[32], GV: GLIDING VERTEX[33]
RRN ROTATED RETINANET[34], RFC: ROTATED FCOS[35], ORC: ORIENTED RCNN[36]
* INDICATES THAT FPN IN THE DETECTOR IS REPLACED WITH BAFPN, ** INDICATES THAT THE ORDER OF THE TWO BRANCHES IN BAFPN IS SWAPPED.

| Method | Backbone | PL | BD | BR | GTF | SV | LV | SH | TC | BC | ST | SBF | RA | HA | SP | HC | CC |
|---|---|---|---|---|---|---|---|---|---|---|---|---|---|---|---|---|---|
| RFR | ResNet50 | 72.27 | 81.41 | 49.23 | 66.71 | 51.81 | 71.48 | 80.19 | 90.82 | 79.08 | 67.95 | 54.15 | 72.28 | 65.40 | 64.79 | 56.61 | **9.06** |
| RFR* | ResNet50 | **72.26** | **82.12** | **49.08** | **68.07** | **51.91** | **71.92** | **80.24** | **90.88** | 78.22 | **68.45** | **55.05** | 71.50 | **66.39** | **66.23** | 55.04 | 7.51 |
| RTS | ResNet50 | 72.21 | 82.25 | 52.89 | **71.68** | 52.53 | 75.70 | **80.97** | **90.89** | 79.17 | 69.56 | 57.63 | 72.54 | 68.26 | 65.14 | 56.16 | 1.67 |
| RTS* | ResNet50 | 72.17 | **83.07** | **54.38** | 71.33 | **52.54** | **76.55** | 80.96 | 90.88 | **79.48** | **69.60** | **61.35** | **72.83** | **74.89** | **71.30** | **58.03** | **13.22** |
| GV | ResNet50 | **72.20** | 76.82 | 46.47 | 66.70 | 51.80 | 72.90 | 80.63 | **90.87** | **77.47** | 68.03 | 52.94 | **72.72** | 65.22 | **65.31** | **51.78** | 5.19 |
| GV* | ResNet50 | 72.15 | **80.33** | **51.83** | **68.92** | **51.84** | **74.15** | **80.69** | **90.87** | 77.38 | **68.29** | **57.46** | 70.76 | **66.84** | 64.93 | 49.57 | **5.55** |
| RRN | ResNet50 | 76.36 | 77.41 | 39.99 | 67.51 | 45.85 | 58.82 | 77.11 | 90.85 | 75.57 | 67.11 | 47.22 | 71.49 | 52.82 | 62.73 | 41.10 | 0.0 |
| RRN* | ResNet50 | **74.70** | **81.06** | **41.60** | **67.79** | **46.17** | **59.58** | **77.20** | **90.87** | **77.34** | **69.42** | **50.90** | **71.55** | **57.69** | **62.97** | **45.27** | 0.0 |
| RFC | ResNet50 | 78.88 | 71.17 | 45.32 | 57.28 | 54.18 | 71.69 | 79.95 | **90.87** | 75.63 | 73.21 | 53.10 | 69.01 | 60.50 | 63.55 | 42.78 | 0.87 |
| RFC* | ResNet50 | **79.31** | **74.30** | **49.33** | **60.59** | **56.63** | **73.49** | **83.01** | 90.86 | **78.13** | **74.14** | **58.62** | **68.38** | **64.34** | **67.62** | **47.41** | **11.75** |
| ORC | ResNet101 | 79.60 | 81.48 | 52.51 | 71.58 | **52.27** | 76.22 | **81.05** | 90.87 | 77.66 | 67.96 | 57.46 | 70.73 | 68.04 | 65.46 | 50.02 | 7.02 |
| ORC* | ResNet101 | **79.72** | **82.34** | **54.54** | **73.87** | 52.22 | **76.60** | 80.96 | **90.88** | **79.30** | **69.13** | **58.72** | **72.51** | **75.57** | **67.58** | **56.31** | **18.81** |
| ORC | ResNet50 | 79.74 | 81.95 | 52.90 | 70.69 | 52.16 | 75.95 | 80.90 | 90.83 | 78.81 | 68.08 | 56.93 | 72.17 | 67.10 | 64.89 | 60.58 | 14.23 |
| ORC-N | ResNet50 | 79.94 | 78.30 | **54.42** | 68.84 | 52.22 | 76.16 | **87.23** | 90.84 | 78.47 | 68.15 | 55.83 | **80.60** | 67.34 | **66.10** | 48.19 | 12.28 |
| ORC-P | ResNet50 | 79.49 | 81.44 | 52.21 | 71.02 | 52.22 | 76.30 | 80.98 | **90.86** | **81.18** | 68.46 | 59.87 | 72.26 | 67.43 | 64.26 | 56.83 | 17.25 |
| ORC*-R | ResNet50 | 79.56 | 81.83 | 52.95 | **71.94** | 52.15 | 76.27 | 80.93 | **90.86** | 78.43 | 68.26 | 60.14 | 72.85 | 67.86 | 65.61 | **62.50** | 17.60 |
| ORC* | ResNet50 | **80.09** | **83.04** | 54.39 | 71.61 | **52.37** | **76.54** | 87.14 | 90.85 | 79.15 | **68.93** | **62.66** | 73.23 | **68.11** | 64.96 | 59.60 | 17.34 |

### C. Main Results

The baseline method in this paper use the classic two-stage detector Oriented RCNN with ResNet50[39] as the backbone network and FPN as the neck network. We summarize the experimental results on the DOTAv1.5 dataset in Tables 1 and 2. Among them, methods with an asterisk (*) in the abbreviation indicate that FPN is replaced by BAFPN, the suffix "R" indicates that the order of the two paths in BAFPN is reversed, the suffix "N" indicates that FPN is replaced by NAS-FPN, and the suffix "P" indicates that FPN is replaced by PAFPN.

After replacing the FPN in Oriented RCNN with BAFPN, we observed that the AP75 increased from 40.22% to 41.9%, AP50 increased from 66.67% to 68.12%, and mAP increased from 39.6% to 40.84%. Then, we replaced the FPN in another classic two-stage detector, Rotated FasterRCNN, with BAFPN, which improved the detector's performance by 3%. We also conducted experiments on two one-stage detectors, Rotated RetinaNet and Rotated FCOS. Both of these detectors discarded the first-layer features from the Backbone output, which prevented BAFPN from acquiring target location information from the highest resolution features for aligning deeper features. Nevertheless, BAFPN still improves the mAP, AP50, and AP75 of anchor-based Rotated RetinaNet by 1.36%, 1.38%, and

1.61%, respectively, and increases these metrics by 3 AP for the anchor-free Rotated FCOS.

The performance improvement brought by BAFPN to Rotated FCOS is highly significant. We hypothesize that anchor-free methods are more susceptible to the effects of misaligned fused features. Deep features contain richer semantic information about the target; however, if these deep features undergo positional shifts or shape distortions and are fused into shallow layers, they may interfere with the Center-ness branch in the FCOS detector, which determines the central points of targets. Misjudged central points are often classified as negative samples because they fall outside the positive sample range defined by the ground truth (GT) boxes. This misclassification reduces the number of positive samples, thereby lowering the chances of detecting the target. Additionally, if regions containing target semantic information are classified as negative samples, the training process becomes more challenging, ultimately leading to a significant decline in detector performance. In contrast, anchor-based methods usually group both the original and misaligned features into the same anchor box. Consequently, misaligned features still contribute to the localization and classification of targets, making anchor-based methods less affected by misaligned fused features compared to anchor-free methods. BAFPN has the ability to align misaligned and distorted features on a global scale, enabling the Center-ness branch to make more precise judgments about the target's center points. As a result, BAFPN achieves a more substantial performance improvement for the anchor-free Rotated FCOS. Similarly, BAFPN boosts the AP75 of the anchor-free Gliding Vertex detector, which is based on four corner-point localization, by 1.97 percentage points, further validating BAFPN's effectiveness for anchor-free detectors. BAFPN can even improve the AP75 of the powerful RoiTransformer by 1 AP.

TABLE II. COMPREHENSIVE ACCURACY COMPARISON

| Method | Backbone | mAP(%) | AP50(%) | AP75(%) |
|---|---|---|---|---|
| RFR | ResNet50 | 35.98 | 64.58 | 34.79 |
| RFR* | ResNet50 | **36.73** | **64.68** | **35.80** |
| RTS | ResNet50 | 40.14 | 65.58 | 41.59 |
| RTS* | ResNet50 | **41.12** | **67.66** | **42.53** |
| GV | ResNet50 | 33.93 | 63.57 | 31.39 |
| GV* | ResNet50 | **35.18** | **64.47** | **33.36** |
| RRN | ResNet50 | 34.59 | 59.50 | 34.67 |
| RRN* | ResNet50 | **35.95** | **60.88** | **36.28** |
| RFC | ResNet50 | 34.26 | 61.75 | 32.85 |
| RFC* | ResNet50 | **37.22** | **64.87** | **35.98** |
| ORC | ResNet101 | 40.27 | 65.62 | 42.15 |
| ORC* | ResNet101 | **41.30** | **68.07** | **43.02** |
| ORC | ResNet50 | 39.42 | 66.74 | 40.22 |
| ORC-N | ResNet50 | 39.49 | 65.93 | 41.15 |
| ORC-P | ResNet50 | 40.00 | 67.00 | 41.10 |
| ORC*-R | ResNet50 | 40.48 | 67.48 | 41.60 |
| ORC* | ResNe50 | **40.84** | **68.12** | **41.90** |

We replaced the backbone of Oriented RCNN with the more powerful ResNet101 and conducted 24 epochs of sufficient training. We found that BAFPN still improved the mAP and AP75 of the detector by 1 AP and AP50 by over 2 AP. Compared to the shallower ResNet50, ResNet101 has more powerful feature extraction capabilities, but the image will pass through more convolutional layers during feature extraction, exacerbating issues such as shape distortion and position misalignment in high-level features. BAFPN aligns features' spatial positions on a global scale and can mitigate the position misalignment of deep features in the Backbone, thereby improving target localization performance. As shown in Tables 1 and 2, BAFPN effectively improves model performance across different backbones, detectors, and detection methods, proving BAFPN's strong generalization ability and robustness.

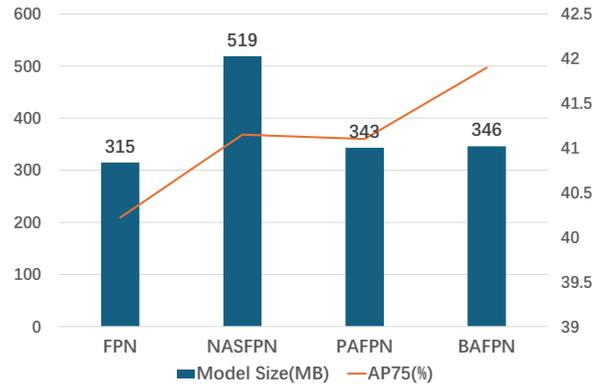

Fig. 6. Model size vs. AP75 comparison

To verify BAFPN's advantage in localization performance over other advanced FPN variants, we replaced the FPN in Oriented RCNN with two advanced plug-and-play FPNs: PAFPN [38] and NAS-FPN, and compared them with the BAFPN detector. The NAS-FPN [24] stack was set to 7. As observed in Tables 1 and 2, BAFPN outperforms both FPN variants in prediction accuracy across most classes. As shown in Figure 6, NAS-FPN increases the model size by 65% but only improves the AP75 by 0.93%. PAFPN increases the model size by 9%, with mAP, AP50, and AP75 improving by 0.58%, 0.26%, and 0.87%, respectively. After the top-down fusion, PAFPN adds a bottom-up information propagation path, but this path is only used to supplement high-level features with additional information. It does not address the position misalignment issue in deep features, and the information passed through this path may come from misaligned fused features. In contrast, the bottom-up feature alignment path in BAFPN, combined with SPAM, effectively resolves this issue. Therefore, although BAFPN's model size is only 3 MB larger than PAFPN's, its performance improvement for the detector is more significant. We believe that, under more stringent target localization requirements, BAFPN has a greater advantage compared to the other two FPN variants.

*D. Ablation Study*

To explore the contribution of each component to the performance of BAFPN, we gradually added the individual modules of BAFPN to the baseline model. In the ablation experiments, GALM was replaced with a single 1×1 convolution, SPAM was replaced with a single 3 × 3 convolution, and OEAM was replaced with nearest-neighbor downsampling. We summarized the experimental results in Table 3, where the circle (O) in the SPAM column indicates that STDDS was replaced with a stride 2 3×3 convolution.

TABLE III. ABLATION STUDY RESULTS

| GALM | SPAM | SEAM | mAP(%) | AP50(%) | AP75(%) |
|---|---|---|---|---|---|
|  |  |  | 39.42 | 66.74 | 40.22 |
| √ |  |  | 39.98 | 66.98 | 41.00 |
|  | √ |  | 40.22 | 66.48 | 41.49 |
|  | O |  | 39.96 | 66.72 | 41.28 |
|  |  | √ | 39.90 | 67.20 | 40.78 |
| √ | √ |  | 40.45 | 67.15 | 41.88 |
| √ |  | √ | 40.19 | 67.25 | 40.88 |
|  | √ | √ | 40.48 | 67.54 | 41.49 |
| √ | √ | √ | **40.82** | **68.09** | **41.90** |

From Table 3, it can be observed that each module effectively improved the performance of the baseline model. Among them, SPAM increased the model's AP75 by 1.27%. Combined with the prepositive bottom-up information propagation path, the performance improvement brought by SPAM alone surpassed that of PAFPN. This shows that SPAM can effectively align the spatial position of the target in the deep feature before the feature misalignment fusion occurs. SEAM improved all of the model's metrics by nearly 0.5 AP, while GALM boosted AP75 by 0.78%. These results validate the effectiveness of SEAM and GALM. We will later explain how SEAM reduces the semantic gap between two feature maps while maintaining feature diversity. When any two modules are combined, the detector's performance is further improved. When all modules are combined, the detector's performance reaches its optimal level.

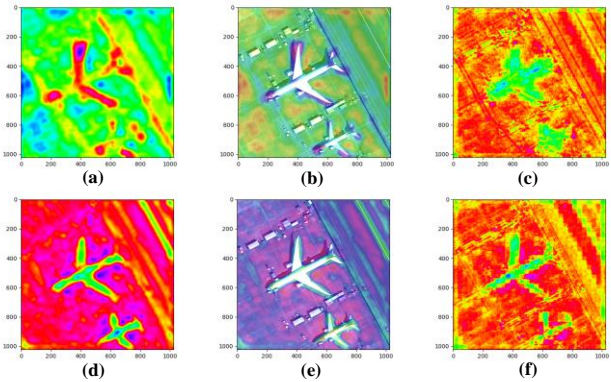

Fig. 7. Feature Distortion and Repair in FPN

To verify the necessity of global feature alignment in improving the detector's localization accuracy, we retained only the SPAM in BAFPN and replaced STDDS with a 3×3 convolution with a stride of 2. In this case, BAFPN becomes equivalent to a PAFPN [38], where the two paths are reversed and a simple DCN feature alignment module is added. Compared to the detector containing only SPAM, the model performance slightly declined but still outperformed PAFPN, which demonstrates the necessity of global feature alignment for improving the detector's localization accuracy, while also validating the effectiveness of STDDS. With other model parameters unchanged, the number of parameters in STDDS is only 46% of that in a 3×3 convolution. Next, we swapped the order of the two branches in BAFPN, which is equivalent to integrating BAFPN's components into PAFPN. We found that the detector's performance decreased slightly compared to the BAFPN detector. This suggests that when top-down feature fusion is performed first, some distorted features may be fused into incorrect positions, activating incorrect features, which interferes with the bottom-up feature alignment process and ultimately impacts the model's detection performance. However, SPAM still provides reliable complement of location information for high-level features, allowing the detector's performance to remain significantly superior to both the baseline model and the baseline model integrated with PAFPN.

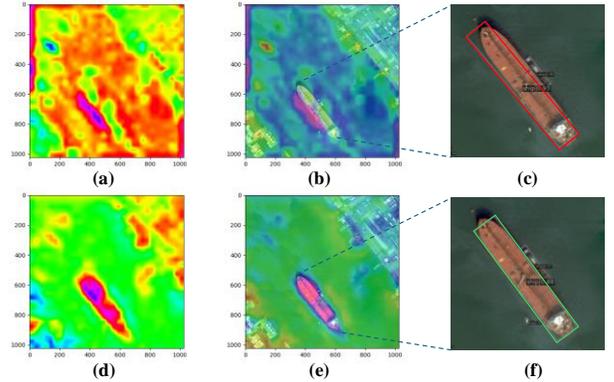

Fig. 8. Feature Shifting and Correction in FPN

To visually demonstrate the advantages of BAFPN in feature alignment, we compared the feature visualization of FPN and BAFPN at the same level. As shown in Figures 7a and 7b, the features extracted from FPN exhibit significant distortion in the shape of the targets, which leads to confusion in the low-level feature representations after cross-scale fusion (Figure 7c). In contrast, the features extracted by BAFPN exhibit regular shapes (Figures 7d and 7e), which closely match the shape of the targets in the original image. After top-down cross-scale fusion, the feature representation remains clear (Figure 7f).

As shown in Figures 8a and 8b, the features extracted from FPN exhibit significant positional shifts, which result in the detector failing to accurately localize the targets (Figure 8c). In contrast, the features extracted from BAFPN more closely align with the target positions in the original image and are expressed more clearly (Figures 8d and 8e), leading to more precise localization of the target by the detector (Figure 8f).

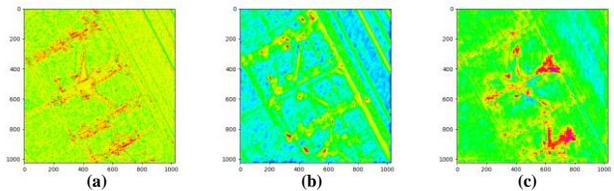

Fig. 9. Semantic Alignment Feature Visualization

Figure 9b shows the second-level features input into BAFPN, which focus on representing the overall information of small targets, while Figure 9a shows the first-level features input into BAFPN, which emphasize the contour information of the targets. It can be observed that there is a significant semantic gap between features from different levels. The global mask (Figure 9c) reduces the semantic gap between the two layers by assigning higher weight to the contour information in the deep features, while also maintaining the high significance of the overall small target features to ensure the diversity of feature representation.

Figures 10a and 10b show the third-level features visualized in FPN, where the target features within the black boxes exhibit insufficient significance, even being confused with surrounding noise. This directly leads to the detector missing the aforementioned targets (Figure 10g). In contrast, the third-level features visualized in BAFPN are clearer (Figures 10c and 10d), but the target features within the black boxes still lack sufficient significance. However, the positional information extracted from lower-level features by STDDS enhances the significance of this region (Figures 10e and 10f). When this positional information is fused with higher-level features, it supplements critical location details, enabling the detector to successfully detect the two targets that were missed in Figure 10g (Figure 10h).

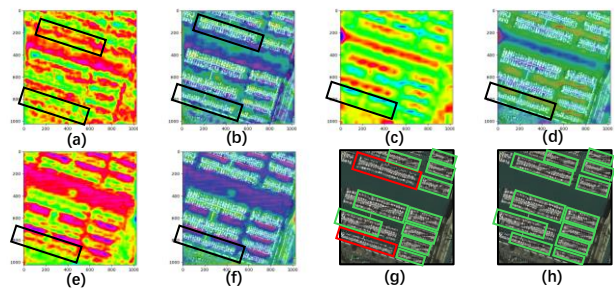

**Fig. 10. Visualization of Space-to-Depth Downsampled Features**

## V. CONCLUSION

This paper provides a detailed analysis of the limitations of FPN and its various variants in remote sensing image object detection, and offers a comprehensive improvement by proposing a novel BAFPN. By integrating SPAM, SEAM, and a novel preceding bottom-up information propagation path, BAFPN aligns features in both directions, addressing the issues of feature misalignment and shape distortion at the global scale. Furthermore, GALM mitigates the information loss caused by the lateral connections of 1×1 convolutions, significantly enhancing the localization accuracy of various detectors in remote sensing image object detection. Through extensive experiments and feature visualizations on the highly challenging DOTAv1.5 dataset, we demonstrate the effectiveness of BAFPN.

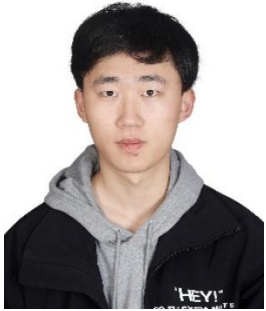

Jiakun Li: Bachelor of Engineering, graduated from Shandong University of Science and Technology, currently pursuing a master's degree at Harbin Engineering University, with a research focus on directed object detection in remote sensing images

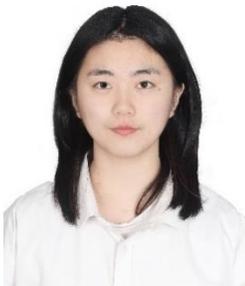

Qingqing Wang: Bachelor of Engineering, graduated from Shandong University of Science and Technology, currently pursuing a master's degree at Harbin Engineering University. Research focuses on low-light image processing, image segmentation, and object detection.

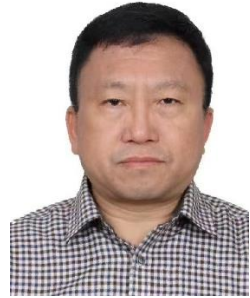

Hongbin Dong: Corresponding author, Doctor of Engineering, Professor, Master's and Doctoral Supervisor, Senior Member of the Computer Society, IEEE Member, and Member of the Artificial Intelligence and Pattern Recognition Professional Committee of the Chinese Computer Society. In recent years, I have visited renowned universities and research institutions in the United States, Spain, Singapore, and Hong Kong for academic exchanges. At present, I am mainly engaged in research on computational intelligence, machine learning, data mining, and multi-agent systems.

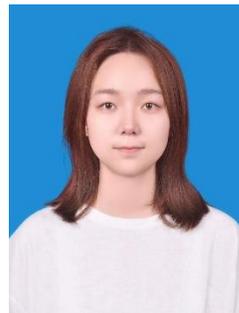

Kexin Li: Currently studying at Shandong Jianzhu University, with research interests in computer vision and generative models.